\documentclass{article}

    \PassOptionsToPackage{numbers, compress}{natbib}

\usepackage[preprint]{neurips_2023}
\usepackage{graphicx} 
\usepackage{amsmath}

\usepackage[utf8]{inputenc} 
\usepackage[T1]{fontenc}    
\usepackage{hyperref}       
\usepackage{url}            
\usepackage{booktabs}       
\usepackage{amsfonts}       
\usepackage{nicefrac}       
\usepackage{microtype}      
\usepackage{xcolor}         

\title{Reflection-Tuning:\\Data Recycling Improves LLM Instruction-Tuning}

\author{
    Ming Li\textsuperscript{\rm 1}, Lichang Chen\textsuperscript{\rm 1}, Jiuhai Chen\textsuperscript{\rm 1}, Shwai He\textsuperscript{\rm 1}, Heng Huang\textsuperscript{\rm 1}, Jiuxiang Gu\textsuperscript{\rm 2}, Tianyi Zhou\textsuperscript{\rm 1}\\
    \textsuperscript{\rm 1}University of Maryland~~~~
    \textsuperscript{\rm 2}Adobe Research\\
    \texttt{\{minglii, bobchen, tianyi\}@umd.edu} \\
}

\begin{document}

\maketitle

\begin{abstract}
Recent advancements in Large Language Models (LLMs) have expanded the horizons of natural language understanding and generation. Notably, the output control and alignment with the input of LLMs can be refined through instruction tuning. 
However, as highlighted in several studies, low-quality data in the training set are usually detrimental to instruction tuning, resulting in inconsistent or even misleading LLM outputs.  
We propose a novel method, termed ``reflection-tuning,'' which addresses the problem by self-improvement and judging capabilities of LLMs. This approach utilizes an oracle LLM to recycle the original training data by introspecting and enhancing the quality of instructions and responses in the data. Extensive experiments on widely used evaluation benchmarks show that LLMs trained with our recycled data outperform those trained with existing datasets in various benchmarks. Codes, data, and models are available in \url{https://github.com/MingLiiii/Reflection_Tuning}.

\end{abstract}

\section{Introduction}

Recently, the emergence and rapid advancement of Large Language Models (LLMs) \cite{touvron2023llama, touvron2023llama2, penedo2023refinedweb, Scao2022BLOOMA1} have pushed the boundaries of natural language understanding and generation. These models have been applied to a variety of applications \cite{zhao2023survey, yang2023harnessing}, from content generation to answering complex questions. A salient feature of LLMs is their potential to follow instructions given to them, a characteristic that has been harnessed to fine-tune and control their outputs. This process, commonly referred to as instruction tuning  \cite{wei2022finetuned, Longpre2023TheFC, chen2023alpagasus, mishra2021cross, Chung2022ScalingIL, zhang2023instruction}, holds immense promise for customizing LLMs to specific tasks or preferences.

However, instruction tuning is susceptible to the quality of training data. Introducing suboptimal data into the training process can have a cascade of adverse effects. Within the ambit of natural language generation, empirical research delineates that both the integrity and the homogeneity of training data critically modulate the fluency, pertinence, and precision of the generated linguistic content \cite{bender-friedman-2018-data, dodge-etal-2019-show, gebru2021datasheets}. Datasets exhibiting inconsistencies or subpar quality can precipitate models to engender erratic, prejudiced, or even specious outputs, thereby attenuating their dependability and applicability. Analogous issues permeate instruction-tuning environments. 
Recent research \cite{yan2023virtual, shu2023exploitability} underscores that even a minuscule fraction of skewed virtual prompts can severely impinge upon a model's operational efficacy, manifesting the susceptibility of large language models (LLMs) to inferior data. 
On the other hand, ALPAGASUS \cite{chen2023alpagasus} and Cherry LLM \cite{li2023quantity} demonstrate that LLMs can achieve enhanced performance metrics by leveraging a select subset of high-quality data.

To address this identified challenge, we introduce a novel method engineered to enhance the quality of extant instruction-tuning datasets autonomously. Drawing inspiration from the evaluative proficiencies of LLMs \cite{zheng2023judging, vicuna2023, alpaca_eval} and contemporary paradigms in self-enhancement \cite{self-improve, pan2023automatically}, our approach hinges on employing an oracle model to introspectively assess and improve the current dataset against specific criteria. This process of data refinement, which we term ``reflection-tuning'', constitutes a potent and efficacious mechanism to bolster the quality of instruction-tuning data. Crucially, this approach obviates the need for supplementary model training and boasts universal adaptability to diverse instruction-response pair architectures. While analogous methodologies have been broached in recent self-alignment literature \cite{self-improve, chen2023teaching, bai2022constitutional} – typified by their application of the model for its own enhancement or in aligning model outputs with preconceived critiques – our contribution is pioneering in integrating the reflection and modification paradigm to both instruction and response dimensions, thereby facilitating the genesis of superior instruction-tuning datasets.

Our extensive experiments include comprehensive evaluations of the models trained with reflection-tuning, including the instruction-following evaluations, e.g., Alpaca-Eval, some human-instruction test sets, and benchmarks. Since GPT-4 demonstrates higher agreement with human preferences than agreements between humans~\cite{zheng2023judging}, we utilize it as our judge for our main instruction-following evaluations. In the comparison with the models trained with the original datasets, e.g., Alpaca \cite{alpaca}, WizardLM \cite{xu2023wizardlm}, our reflection-tuned models achieve much better performance. Specifically, our recycled WizardLM 7B model achieves the highest win rate among other open-source 7B models in the Alpaca-Eval leaderboard. Moreover, Our recycled Alpaca achieves a win rate of $88.75\%$ and our recycled WizardLM achieves a win rate of $81.25\%$ on the Vicuna \cite{vicuna2023} test set with the same number of training data and model size.

\section{Related Work}
\paragraph{Instruction Tuning of LLMs.} The overarching goal of our work is to enhance the model's instruction-following capability, which is consistent with the previous works~\cite{Chung2022ScalingIL, Longpre2023TheFC, openai2023gpt4}. It is discovered that the cross-task generalization ability of LLMs could be enhanced by fine-tuning on NLP datasets which are structured with instruction-response pairs~\cite{mishra2021cross, wei2021finetuned}. More recent works~\cite{NEURIPS2022_b1efde53, bai2022training} have expanded instruction tuning to include open-ended generation tasks, which exhibit enhanced handling of complex human instructions.

\paragraph{High-quality data generation.} Our method also targets generating better instruction tuning data~\cite{wang-etal-2023-self-instruct, peng2023instruction, xu2023wizardlm}, but it is orthogonal to the previous work since any kind of instruction-response pairs can be further reflected and improved by our method. Recent works either curate the instruction tuning datasets by human labors, e.g., Dolly~\cite{DatabricksBlog2023DollyV2}, Longpre~\cite{Longpre2023TheFC} or distill the responses from SOTA LLMs like GPT4~\cite{openai2023gpt4}, e.g., Alpaca~\cite{alpaca}, Alpaca-GPT4~\cite{peng2023instruction}, Vicuna~\cite{vicuna2023}, Koala~\cite{vu2023koala}. There is also some exploration of making the instructions more difficult through the evolution~\cite{xu2023wizardlm}, which achieves incredible performance on Alpaca-Eval~\cite{alpaca_eval}. Different from them, our method could be treated as a useful posthoc tool, which can further enhance the quality of the instruction tuning data.

\paragraph{LLM self-alignment.} Our study contributes to the expanding body of self-alignment~\cite{principle, self-improve}, i.e., it proves the self-check and self-refine ability of the LLMs. Constitutional-AI~\cite{bai2022constitutional} first introduces the idea of using the feedback of the AI itself as the preference data to optimize the objectives of helpfulness and harmlessness. Recent works~\cite{chen2023teaching, li2023self, lee2023rlaif} show that LLMs can generate useful signals for debugging, filtering, and finetuning with RL. These works inspire our study prompting the ChatGPT to self-reflect its own generated responses and then self-revise.

\section{Methodology}

\subsection{Preliminaries}

Initially, we elucidate and formalize extant methodologies that leverage large language models for instruction-tuning. Let $f_\theta$ denote the pre-trained LLM, e.g., Llama, with parameters $\theta$ and $g$ the oracle LLM, e.g., ChatGPT. We use other lowercase letters $x, y, z, c,..$ to denote the text segments, which could be phrases or sentences, and each token in $x$ is denoted as $x[i]$. We use uppercase letters $D, ..$ to denote the collection of language sequences or datasets, and $D_0$ represents the initial base dataset. Since both $f_\theta$ and $g$ are in auto-regressive manners, a sequence $x = (x[1],...,x[n])$ can be further denoted as $f_\theta(x) = \prod_{i=1}^{n} f(x[i] | x[1,...,i])$.

In the instruction-following setting, there will be a mapping function that turns the original raw instruction $x$ into the desirable format and requests models for a response $y$. For simplicity, we directly notate this process as $y \sim f(y|x)$. And the loss function for instruction-tuning can be denoted as $L = -\frac{1}{n}\sum_{i=1}^{n} \log f_\theta(y|x)$ where $n$ is the length of response $y$.

\subsection{Reflection-Tuning}

There are two main phases in our method, instruction reflection and response reflection. 
Based on the intuition that students who reflect on the answers usually get higher scores since they can find the errors and make some reasonable changes through the reflection process, and astonished by the self-improvement \cite{self-improve, pan2023automatically} and judging \cite{zheng2023judging, vicuna2023, alpaca_eval} capability of LLMs, we propose a reflection method for improving the quality of instruction-response pairs. Given the initial base dataset, we are motivated to generate a high-quality version of each data point with an oracle model, ChatGPT for instance. However, a common problem with using LLMs as judges is the failure to obtain diverse results. To overcome this potential problem, inspired by Chain-of-Thought and Tree-of-Thought prompting \cite{wei2023chainofthought, yao2023tree}, we further define several specific criteria $\{c_1, ..., c_k\}$ for the oracle model to follow, and respond to those specific criteria with critical responses $\{z_1, ..., z_k\}$, respectively. Then the responses to these criteria can bridge the generation of new instruction-response pairs. 

\subsubsection{Reflection on Instruction}
Specifically, in the instruction reflection phase, the oracle model $g$ is required to reflect on the given instruction-response pair $(x^0, y^0)$ from the original dataset $D^0$ with some specific criteria $\{c_1^{ins}, ..., c_k^{ins}\}$ and then generate a better instruction-response pair $(x^{ins}, y^{ins})$ according to its reflection results. With the criteria given, the oracle model $g$ is able to generate critical responses: 
\begin{equation}
    [ z_1^{ins}, ..., z_k^{ins} ] \sim g(z_1^{ins}, ..., z_k^{ins}|x^0, y^0, c_1^{ins}, ..., c_k^{ins})
\end{equation}
where both original instruction and response are wrapped into the prompt rather than original instruction alone. These critical responses further serve as the guidance (chain of thought) for the generation of the new instruction and response pair: 
\begin{equation}
    [ x^{ins}, y^{ins} ] \sim g( x^{ins}, y^{ins}|x^0, y^0, c_1^{ins}, ..., c_k^{ins}, z_1^{ins}, ..., z_k^{ins})
\end{equation}
where in practice the above process is sampled as a continuous language sequence, and the critical responses would not be decomposed from the whole outputs. 
The criteria used for instruction are ``the Complexity of the Topic'', ``the Level of Detail Required for response'', ``Knowledge Required for response'', ``the Ambiguity of the Instruction'' and whether ``Logical Reasoning or Problem-Solving Involved''.

\subsubsection{Reflection on Response}
Although both instruction and response are modified, the corresponding response $y^{ins}$ for a given modified instruction $x^{ins}$ is not optimal. Thus another reflection on the response process is further proposed. Similar to the above procedure, a new set of criteria for reflection on response is defined as $\{c_1^{res}, ..., c_m^{res}\}$. The overall process can be noted as: 
\begin{equation}
    y^{res} \sim g(y^{res}|x^{ins}, y^{ins}, c_1^{res}, ..., c_m^{res}, z_1^{res}, ..., z_m^{res})
\end{equation}
where $z_i^{res}$ represents the critical response of $i$th response criteria $c_i^{res}$. After the above process, the instruction and response pair $(x^{ins}, y^{res}))$ is regarded as the recycled data pair which will be used for instruction-tuning of model $f_\theta$. 
The criteria used for instruction are ``Helpfulness'', ``Relevance'', ``Accuracy'', and ``Level of Details''. 

We name the whole above process as a recycling process, which greatly improves the quality of the previous dataset. Then the raw model $f_\theta$ will be trained on the newly generated recycled dataset, and the newly generated models are notated as ``Recycled Models'', eg. Recycled Alpaca. 

\section{Experimental Setup}

\subsection{Base Datasets}

The Alpaca dataset \cite{alpaca}, sourced from Stanford University, offers $52,002$ instruction-following samples. Developed via the self-instruct paradigm \cite{wang-etal-2023-self-instruct}, it leveraged the capabilities of the text-davinci-003 model. This dataset, while a pioneering attempt in instruction tuning for the LLaMA model, raised concerns about data quality owing to its reliance on the text-davinci-003 model.

On the other hand, the WizardLM dataset \cite{xu2023wizardlm}, which employs the sophisticated Evol-Instruct algorithm, is a refined collection encompassing a total of $250,000$ instruction samples. Two primary evolutionary trajectories, namely "In-depth Evolving" and "In-breadth Evolving", are introduced within this dataset. These trajectories are specifically designed to allow a base instruction to progress either in terms of intricate details or in its overall scope. To enhance data fidelity, ChatGPT has been meticulously integrated during the refinement process. From this extensive dataset, we predominantly focused on the WizardLM-7b subset, comprising $70,000$ samples.
We test our method on both of these two datasets to verify the effectiveness of our method. 

\subsection{Implementation Details}

Rooted in the Llama2-7b pre-trained model \cite{touvron2023llama2}, we utilize the prompt and code base from Vicuna and flash attention while the overall training arguments are aligned with protocols from Alpaca and WizardLM datasets. The Adam optimizer \cite{kingma2017adam}, with a $2\times10^{-5}$ learning rate and a batch size of $128$, steers the training across three epochs with a max length of $2048$. The warmup rate is set to $0.03$. 

\subsection{Evaluation Metric}

\subsubsection{Pari-wise comparison}

The task of quantitatively evaluating the instruction-adherence efficacy of LLMs presents considerable challenges. Despite a wealth of research endeavoring to design automated evaluation metrics for LLMs \cite{chang2023survey}, the gold standard remains subjective human evaluation. However, such manual assessments are not only resource-intensive but are also susceptible to inherent human biases. 

Incorporating methodologies from cutting-edge LLM evaluations \cite{zheng2023judging, vicuna2023, alpaca_eval}, we operationalize GPT4 and ChatGPT as evaluation benchmarks. As delineated in \cite{chen2023alpagasus}, models subjected to evaluation are prompted to generate outputs for each instruction in the test corpus. Subsequent to this, an API-driven model, be it GPT4 or ChatGPT, allocates a score to each response. A model's superiority on this dataset hinges on its endorsement by the adjudicating model.

The adjudication phase entails rating each model-generated response on a scale spanning from \(1\) to \(10\), with scores encapsulating facets such as pertinence and precision. To mitigate the positional bias elaborated upon in \cite{ko-etal-2020-look, wang2023large}, model-generated outputs are presented to the adjudicating entity in two distinct sequences and subsequently scored. Hence, a model's dominance is ratified under the following conditions: 
\textbf{Wins:} Exhibits superiority in both sequences or prevails in one while maintaining parity in the alternate sequence.
\textbf{Tie:} Demonstrates parity across both sequences or prevails in one while faltering in the alternate.
\textbf{Loses:} Underperforms in both sequences or maintains parity in one while being eclipsed in the alternate.
This adjudication paradigm underpins our experimental findings.

\subsection{Benchmarks}

Two prominent benchmarking platforms for LLMs are highlighted: the Huggingface Open LLM Leaderboard\footnote{\url{https://huggingface.co/spaces/HuggingFaceH4/open_llm_leaderboard}} and the AlpacaEval Leaderboard\footnote{\url{https://tatsu-lab.github.io/alpaca_eval}}. The Huggingface Open LLM Leaderboard employs the evaluation methodology from \cite{eval-harness}, providing a cohesive framework for assessing generative language model capabilities across a spectrum of evaluation tasks. It focuses on $4$ pivotal benchmarks: ARC \cite{clark2018think}, HellaSwag \cite{zellers-etal-2019-hellaswag}, MMLU \cite{hendrycks2021measuring}, and TruthfulQA \cite{lin-etal-2022-truthfulqa}. Specifically, ARC is a specialized dataset curated for assessing the proficiency of models in answering science questions tailored for grade-school levels. The challenge employs a 25-shot learning paradigm, implying that models are exposed to 25 examples prior to evaluation. HellaSwag is Specifically designed to probe models on their commonsense inference capabilities, which utilizes a 10-shot learning setup, meaning models are trained on 10 sample instances before being tested. MMLU is a comprehensive evaluation suite designed to gauge a model's multitasking learning capability across a diverse range of 57 tasks. These tasks span a myriad of domains including but not limited to elementary mathematics, US history, computer science, and jurisprudence. TruthfulQA is constructed to appraise a model's susceptibility to perpetuating misinformation or falsehoods, which are ubiquitously found online.

On the other hand, the AlpacaEval Leaderboard offers an LLM-centric automatic assessment utilizing the AlpacaFarm \cite{dubois2023alpacafarm} evaluation dataset. It is an automated evaluation mechanism for LLMs that offers efficiency, cost-effectiveness, and reliability. Operating on the AlpacaFarm evaluation dataset, it gauges models' proficiency in adhering to generic user instructions. The generated outputs are juxtaposed against benchmark responses from Davinci003. These benchmarks are subsequently auto-annotated by either GPT-4, Claude, or ChatGPT, leading to the determination of the aforementioned win rates. Empirical evidence suggests that AlpacaEval's alignment with ground truth annotations sourced from human experts is notably high. Furthermore, model rankings on the AlpacaEval leaderboard exhibit a strong correlation with rankings derived from human annotators.

\section{Experimental Results}

\subsection{Pair-wise Comparison}

\begin{figure}[!htbh]
\centering 
\includegraphics[width=1\textwidth]{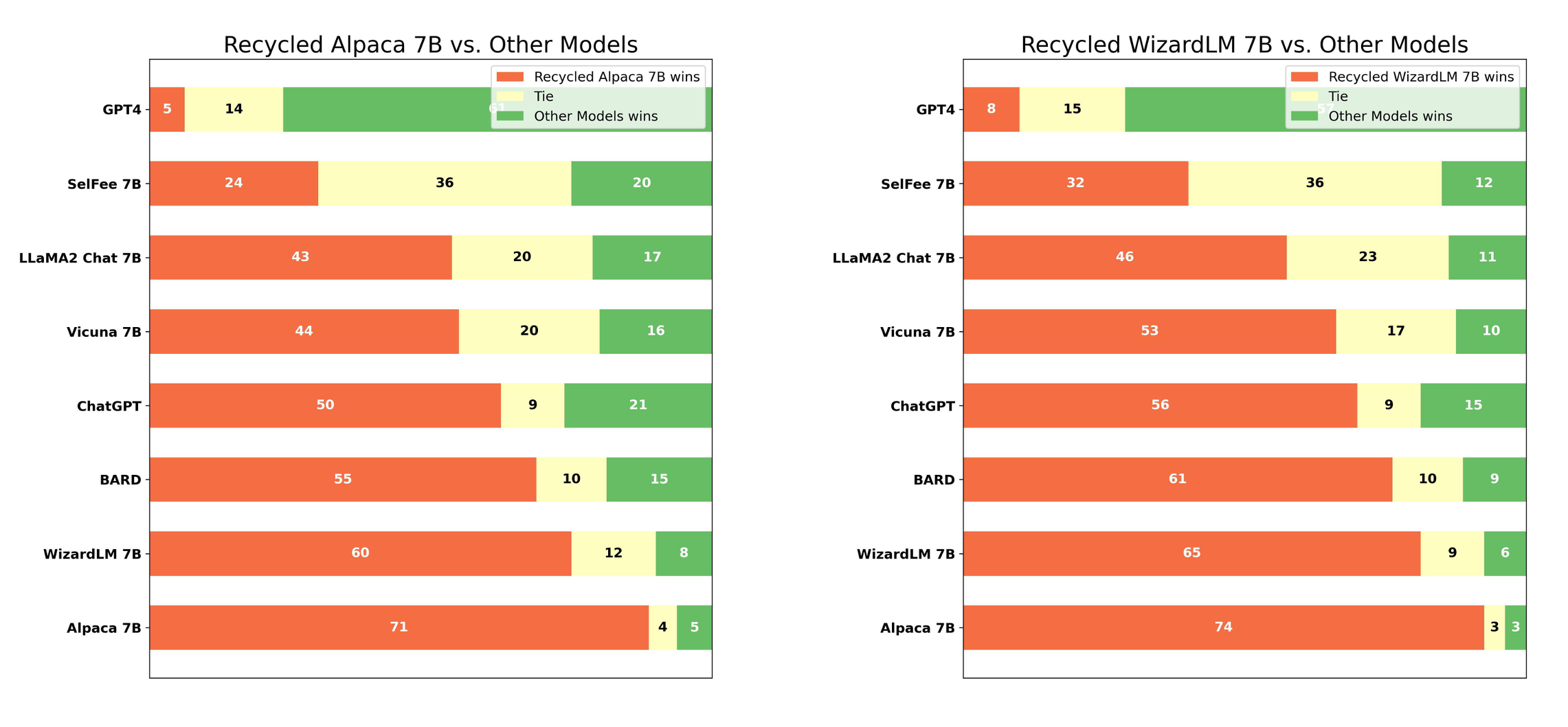} 
\caption{
Comparing our recycled models with other renowned models on the Vicuna evaluation set. On the left list the models that are compared. Each bar represents a comparison between our recycled model and the other model. The red parts represent the number of wins and the green parts represent the number of loses. GPT4 is utilized as the judge. 
} 
\label{pair_compare} 
\end{figure}

As depicted in Figure \ref{pair_compare}, a juxtaposition between our recycled models and other distinguished models is presented. Remarkably, our models exhibit superior performance across the board, with GPT4 being the sole exception, underscoring the efficacy of our methodology. Notably, SelFee \cite{selfee2023} aligns with our motivation in leveraging an oracle model to refine dataset responses while using much more data for training including the Alpaca dataset, the ShareGPT dataset, the FLAN dataset, and extra math and code collections. However, even with much more data used, they overlook the criticality of enhancing the instruction set and neglect the deployment of granular criteria for self-enhancement. This negligence results in their suboptimal performance despite a voluminous training dataset. Importantly, our models, equipped solely with instruction tuning on the Alpaca dataset, surpass several counterparts that employ additional RLHF techniques.

\subsection{Alpaca Eval Leaderboard}

Table \ref{tbl:alpaca_eval} delineates the outcomes on the AlpacaEval Leaderboard. Within this evaluation framework, GPT4 is harnessed as the adjudicating entity, contrasting the responses of the test models against the benchmark set by Davinci003. This comparison provides a direct quantification of a model's capacity for instruction adherence and the intrinsic quality of its output. Notably, our models eclipse the performance of all extant 7B open-source counterparts, with the sole exception being Xwin-LM \cite{xwin-lm} whose training data is unknown and extra RLHF is implemented. Remarkably, our models even surpass some of the models with a larger parameter count. The eminent positioning of our models on this leaderboard underscores the superior caliber of the responses they generate. 

\begin{table}[h]
\centering
\scalebox{0.95}{
\begin{tabular}{l|cccccc}
\hline
\textbf{Model} & \textbf{Win Rate} & \textbf{Standard Error} & \textbf{Wins} & \textbf{Draws} & \textbf{Avg Length} \\
\hline
GPT4 \cite{openai2023gpt4} & 95.28 & 0.72 & 761 & 12 & 1365 \\
Claude 2 & 91.36 & 0.99 & 734 & 1 & 1069 \\
ChatGPT & 89.37 & 1.08 & 716 & 5 & 827 \\
XwinLM 7b V0.1 \cite{xwin-lm} & 87.83 & - & - & - & 1894 \\
\textbf{Recycled WizardLM 7B (ours)} & 78.88 & 1.44 & 635 & 0 & 1494 \\
\textbf{Recycled Alpaca 7B (ours)} & 76.99 & 1.49 & 619 & 0 & 1397 \\
Vicuna 7B v1.3 \cite{vicuna2023} & 76.84 & 1.49 & 614 & 3 & 1110 \\
WizardLM 13B \cite{xu2023wizardlm}& 75.31 & 1.51 & 601 & 9 & 985 \\
airoboros 65B & 73.91 & 1.53 & 587 & 16 & 1512 \\
Guanaco 65B \cite{dettmers2023qlora}& 71.80 & 1.59 & 578 & 0 & 1249 \\
LLaMA2 Chat 7B \cite{touvron2023llama2}& 71.37 & 1.59 & 574 & 1 & 1479 \\
Baize-v2 13B \cite{xu2023baize}& 66.96 & 1.66 & 538 & 2 & 930 \\
Guanaco 33B \cite{dettmers2023qlora}& 65.96 & 1.67 & 531 & 0 & 1311 \\
Vicuna 7B \cite{vicuna2023} & 64.41 & 1.69 & 517 & 3 & 1044 \\
Davinci003 & 50.00 & 0.00 & 0 & 805 & 307 \\
Guanaco 7B \cite{dettmers2023qlora}& 46.58 & 1.76 & 374 & 2 & 1364 \\
Alpaca 7B \cite{alpaca} & 26.46 & 1.54 & 205 & 16 & 396 \\
\hline
\end{tabular}
}
\caption{The comparison of performance on AlpacaEval Leaderboard. }
\label{tbl:alpaca_eval}
\end{table}

\subsection{Open LLM Leaderboard}

Table \ref{tbl:open} showcases the performance comparison on the Huggingface Open LLM Leaderboard with some related models. With our Recycle mechanism, our models achieve better average performances across these four representative benchmarks and our results are comparable to llama-2-7b-chat, which is elaborately fine-tuned with extra RLHF.  

\begin{table*}[h]
\centering
\scalebox{1.0}{
\begin{tabular}{l|ccccc}
\hline
& \multicolumn{5}{c}{Huggingface Open LLM Leaderboard}  \\
& Average & ARC & HellaSwag & MMLU & TruthfulQA  \\
\hline
Alpaca 7B \cite{alpaca} & 50.21 & 42.65 & 76.91 & 41.73 & 39.55 \\
WizardLM 7B \cite{xu2023wizardlm} & 54.18 & 51.60 & 77.70 & 42.70 & 44.70  \\
Vicuna 7B v1.3 \cite{vicuna2023} & 55.63 & 50.43 & 76.92 & 48.14 & 47.01 \\
LLaMA2 Chat 7B \cite{touvron2023llama2}& 56.34 & 52.90 & 78.55 & 48.32 & 45.57 \\
\hline
\textbf{Recycled Alpaca 7B (ours)} & 56.18 & 53.92 & 77.68 & 47.55 & 45.55 \\
\textbf{Recycled WizardLM 7B (ours)} & 56.21 & 53.92 & 77.05 & 48.35 & 45.21  \\
\hline
\end{tabular}
}
\caption{
The comparison of performance on Huggingface Open LLM Leaderboard. 
}
\label{tbl:open}
\end{table*}

\section{Discussion}

\subsection{Statistic Analysis}

In the ensuing discourse, we delve into a quantitative juxtaposition of the instruction-response data, pre- and post-application of our recycling methodology, as delineated in Table \ref{tbl:comparison}. Observationally, there's an increase in the average token length of instructions within the Alpaca dataset, whereas a decrement manifests for the WizardLM dataset, epitomizing the method's adept adaptability. The succinctness and elementary nature of the Alpaca dataset's instructions warrant an enhancement in intricacy through our method, thereby elongating their length. Conversely, the pre-existing complexity and intricacy in WizardLM's instructions render our algorithm inclined towards succinctness. Pertaining to the response section, there's a marked propensity of our approach to engender detail-rich textual content, leading to relatively long responses. Moreover, leveraging Sentence-BERT \cite{reimers-gurevych-2019-sentence}, we quantify the coherence metric between instructions and their affiliated responses. It's discernible that our technique invariably fabricates samples with better coherence, signifying a superior alignment between modulated instructions and consequent responses. Additionally, to elucidate the metamorphosis in instructional difficulty, we employ the Instruction-Following Difficulty (IFD) score, as posited by Cherry LLM \cite{li2023quantity}, executed on the nascent pre-trained language model. This score gauges the efficacy of instructions in bolstering response predictions. The consistent ascension in IFD scores lucidly illustrates our instruction's progressive evolution.

\begin{table*}[h]
\centering
\scalebox{0.9}{
\begin{tabular}{l|ccccccc}
\hline
& \multicolumn{7}{c}{Comparison of Different Models}  \\
& Ins. len & Res. len& Ins. ppl & Res. ppl 1 & Res. ppl 2 & Coherent & IFD score  \\
\hline
Original Alpaca 7B& 20.7 & 65.5 & 34.3 & 82.6 & 49.2 & 0.53 & 0.72 \\
Recycled Alpaca 7B& 37.9 & 377.2 & 13.6 & 4.5 & 2.9 & 0.67 & 0.83\\
\hline
Original WizardLM 7B& 123.0 & 348.5 & 12.3 & 17.0 & 7.5 & 0.65 & 0.66\\
Recycled WizardLM 7B& 66.9 & 518.7 & 10.0 & 3.2 & 2.5 & 0.73 & 0.81\\
\hline
\end{tabular}}
\caption{
The comparison of performance for various models with different metrics. ``Ins. len'' and ``Res. len'' represent the average token length of the instructions and responses. ``Ins. ppl'' represents the average perplexity of instructions. ``Res. ppl 1'' and ``Res. ppl 2'' represent response perplexities without or with the context of corresponding instructions. All the perplexity is calculated upon our initial pre-trained model llama2. ``Coherent'' represents the coherent score calculated by SentenceBert. ``IFD score'' represents the instruction-following difficulty score proposed by Cherry LLM \cite{li2023quantity}. 
}
\label{tbl:comparison}
\end{table*}

\subsection{Performances on 13B Models}

We further train a Recycled Alpaca in the 13B version to further validate the efficacy of our method. With only $52$k recycled alpaca data being used for instruction-tuning, our Recycled Alpaca 13B reaches the win rate of $83.42\%$ in the Alpaca Eval leaderboard and reaches an average score of $58.93\%$ on Huggingface Open LLM leaderboard. Considering the small amount of data we used compared with other models, the results are intriguing and satisfactory. We will soon apply our recycled WizardLM data to the 13B model.

\section{Conclusion}

The evolution of Large Language Models has brought forth unparalleled capacities in natural language processing, especially in the domain of instruction tuning. However, the quality of training data remains a pivotal determinant of model performance. In this work, we introduced the reflection-tuning method, an innovative approach to autonomously improve and recycle the quality of instruction-tuning datasets by leveraging the inherent self-improvement capabilities of LLMs. Our method emphasizes a unique reflect-and-recycle mechanism, a first in the domain, applied comprehensively to both instructions and responses. Experimental results affirm the efficacy of reflection-tuning, with models trained using this method consistently outperforming those trained with traditional datasets. This paves the way for more reliable, consistent, and high-performing LLMs in the future, underscoring the importance of high-quality data recycling and innovative methods in the realm of natural language generation.

\bibliography{neurips_2023}

\begin{thebibliography}{10}

\bibitem{bai2022training}
Yuntao Bai, Andy Jones, Kamal Ndousse, Amanda Askell, Anna Chen, Nova DasSarma, Dawn Drain, Stanislav Fort, Deep Ganguli, Tom Henighan, et~al.
\newblock Training a helpful and harmless assistant with reinforcement learning from human feedback.
\newblock {\em arXiv preprint arXiv:2204.05862}, 2022.

\bibitem{bai2022constitutional}
Yuntao Bai, Saurav Kadavath, Sandipan Kundu, Amanda Askell, Jackson Kernion, Andy Jones, Anna Chen, Anna Goldie, Azalia Mirhoseini, Cameron McKinnon, et~al.
\newblock Constitutional ai: Harmlessness from ai feedback.
\newblock {\em arXiv preprint arXiv:2212.08073}, 2022.

\bibitem{bender-friedman-2018-data}
Emily~M. Bender and Batya Friedman.
\newblock Data statements for natural language processing: Toward mitigating system bias and enabling better science.
\newblock {\em Transactions of the Association for Computational Linguistics}, 6:587--604, 2018.

\bibitem{chang2023survey}
Yupeng Chang, Xu~Wang, Jindong Wang, Yuan Wu, Linyi Yang, Kaijie Zhu, Hao Chen, Xiaoyuan Yi, Cunxiang Wang, Yidong Wang, Wei Ye, Yue Zhang, Yi~Chang, Philip~S. Yu, Qiang Yang, and Xing Xie.
\newblock A survey on evaluation of large language models, 2023.

\bibitem{chen2023alpagasus}
Lichang Chen, Shiyang Li, Jun Yan, Hai Wang, Kalpa Gunaratna, Vikas Yadav, Zheng Tang, Vijay Srinivasan, Tianyi Zhou, Heng Huang, and Hongxia Jin.
\newblock Alpagasus: Training a better alpaca with fewer data, 2023.

\bibitem{chen2023teaching}
Xinyun Chen, Maxwell Lin, Nathanael Sch{\"a}rli, and Denny Zhou.
\newblock Teaching large language models to self-debug.
\newblock {\em arXiv preprint arXiv:2304.05128}, 2023.

\bibitem{vicuna2023}
Wei-Lin Chiang, Zhuohan Li, Zi~Lin, Ying Sheng, Zhanghao Wu, Hao Zhang, Lianmin Zheng, Siyuan Zhuang, Yonghao Zhuang, Joseph~E. Gonzalez, Ion Stoica, and Eric~P. Xing.
\newblock Vicuna: An open-source chatbot impressing gpt-4 with 90\%* chatgpt quality, March 2023.

\bibitem{Chung2022ScalingIL}
Hyung~Won Chung, Le~Hou, S.~Longpre, Barret Zoph, Yi~Tay, William Fedus, Eric Li, Xuezhi Wang, Mostafa Dehghani, Siddhartha Brahma, Albert Webson, Shixiang~Shane Gu, Zhuyun Dai, Mirac Suzgun, Xinyun Chen, Aakanksha Chowdhery, Dasha Valter, Sharan Narang, Gaurav Mishra, Adams~Wei Yu, Vincent Zhao, Yanping Huang, Andrew~M. Dai, Hongkun Yu, Slav Petrov, Ed~Huai hsin Chi, Jeff Dean, Jacob Devlin, Adam Roberts, Denny Zhou, Quoc~V. Le, and Jason Wei.
\newblock Scaling instruction-finetuned language models.
\newblock {\em ArXiv}, abs/2210.11416, 2022.

\bibitem{clark2018think}
Peter Clark, Isaac Cowhey, Oren Etzioni, Tushar Khot, Ashish Sabharwal, Carissa Schoenick, and Oyvind Tafjord.
\newblock Think you have solved question answering? try arc, the ai2 reasoning challenge, 2018.

\bibitem{DatabricksBlog2023DollyV2}
Mike Conover, Matt Hayes, Ankit Mathur, Jianwei Xie, Jun Wan, Sam Shah, Ali Ghodsi, Patrick Wendell, Matei Zaharia, and Reynold Xin.
\newblock Free dolly: Introducing the world's first truly open instruction-tuned llm, 2023.

\bibitem{dettmers2023qlora}
Tim Dettmers, Artidoro Pagnoni, Ari Holtzman, and Luke Zettlemoyer.
\newblock Qlora: Efficient finetuning of quantized llms.
\newblock {\em arXiv preprint arXiv:2305.14314}, 2023.

\bibitem{dodge-etal-2019-show}
Jesse Dodge, Suchin Gururangan, Dallas Card, Roy Schwartz, and Noah~A. Smith.
\newblock Show your work: Improved reporting of experimental results.
\newblock In {\em Proceedings of the 2019 Conference on Empirical Methods in Natural Language Processing and the 9th International Joint Conference on Natural Language Processing (EMNLP-IJCNLP)}, pages 2185--2194, Hong Kong, China, November 2019. Association for Computational Linguistics.

\bibitem{dubois2023alpacafarm}
Yann Dubois, Xuechen Li, Rohan Taori, Tianyi Zhang, Ishaan Gulrajani, Jimmy Ba, Carlos Guestrin, Percy Liang, and Tatsunori~B. Hashimoto.
\newblock Alpacafarm: A simulation framework for methods that learn from human feedback, 2023.

\bibitem{eval-harness}
Leo Gao, Jonathan Tow, Stella Biderman, Sid Black, Anthony DiPofi, Charles Foster, Laurence Golding, Jeffrey Hsu, Kyle McDonell, Niklas Muennighoff, Jason Phang, Laria Reynolds, Eric Tang, Anish Thite, Ben Wang, Kevin Wang, and Andy Zou.
\newblock A framework for few-shot language model evaluation, September 2021.

\bibitem{gebru2021datasheets}
Timnit Gebru, Jamie Morgenstern, Briana Vecchione, Jennifer~Wortman Vaughan, Hanna Wallach, Hal Daumé~III au2, and Kate Crawford.
\newblock Datasheets for datasets, 2021.

\bibitem{hendrycks2021measuring}
Dan Hendrycks, Collin Burns, Steven Basart, Andy Zou, Mantas Mazeika, Dawn Song, and Jacob Steinhardt.
\newblock Measuring massive multitask language understanding.
\newblock In {\em International Conference on Learning Representations}, 2021.

\bibitem{self-improve}
Jiaxin Huang, Shixiang~Shane Gu, Le~Hou, Yuexin Wu, Xuezhi Wang, Hongkun Yu, and Jiawei Han.
\newblock Large language models can self-improve.
\newblock {\em arXiv preprint arXiv:2210.11610}, 2022.

\bibitem{kingma2017adam}
Diederik~P. Kingma and Jimmy Ba.
\newblock Adam: A method for stochastic optimization, 2017.

\bibitem{ko-etal-2020-look}
Miyoung Ko, Jinhyuk Lee, Hyunjae Kim, Gangwoo Kim, and Jaewoo Kang.
\newblock Look at the first sentence: Position bias in question answering.
\newblock In {\em Proceedings of the 2020 Conference on Empirical Methods in Natural Language Processing (EMNLP)}, pages 1109--1121, Online, November 2020. Association for Computational Linguistics.

\bibitem{lee2023rlaif}
Harrison Lee, Samrat Phatale, Hassan Mansoor, Kellie Lu, Thomas Mesnard, Colton Bishop, Victor Carbune, and Abhinav Rastogi.
\newblock Rlaif: Scaling reinforcement learning from human feedback with ai feedback.
\newblock {\em arXiv preprint arXiv:2309.00267}, 2023.

\bibitem{li2023quantity}
Ming Li, Yong Zhang, Zhitao Li, Jiuhai Chen, Lichang Chen, Ning Cheng, Jianzong Wang, Tianyi Zhou, and Jing Xiao.
\newblock From quantity to quality: Boosting llm performance with self-guided data selection for instruction tuning, 2023.

\bibitem{li2023self}
Xian Li, Ping Yu, Chunting Zhou, Timo Schick, Luke Zettlemoyer, Omer Levy, Jason Weston, and Mike Lewis.
\newblock Self-alignment with instruction backtranslation.
\newblock {\em arXiv preprint arXiv:2308.06259}, 2023.

\bibitem{alpaca_eval}
Xuechen Li, Tianyi Zhang, Yann Dubois, Rohan Taori, Ishaan Gulrajani, Carlos Guestrin, Percy Liang, and Tatsunori~B. Hashimoto.
\newblock Alpacaeval: An automatic evaluator of instruction-following models.
\newblock \url{https://github.com/tatsu-lab/alpaca_eval}, 2023.

\bibitem{lin-etal-2022-truthfulqa}
Stephanie Lin, Jacob Hilton, and Owain Evans.
\newblock {T}ruthful{QA}: Measuring how models mimic human falsehoods.
\newblock In {\em Proceedings of the 60th Annual Meeting of the Association for Computational Linguistics (Volume 1: Long Papers)}, pages 3214--3252, Dublin, Ireland, May 2022. Association for Computational Linguistics.

\bibitem{Longpre2023TheFC}
S.~Longpre, Le~Hou, Tu~Vu, Albert Webson, Hyung~Won Chung, Yi~Tay, Denny Zhou, Quoc~V. Le, Barret Zoph, Jason Wei, and Adam Roberts.
\newblock The flan collection: Designing data and methods for effective instruction tuning.
\newblock {\em ArXiv}, abs/2301.13688, 2023.

\bibitem{mishra2021cross}
Swaroop Mishra, Daniel Khashabi, Chitta Baral, and Hannaneh Hajishirzi.
\newblock Cross-task generalization via natural language crowdsourcing instructions.
\newblock {\em arXiv preprint arXiv:2104.08773}, 2021.

\bibitem{openai2023gpt4}
OpenAI.
\newblock Gpt-4 technical report, 2023.

\bibitem{NEURIPS2022_b1efde53}
Long Ouyang, Jeffrey Wu, Xu~Jiang, Diogo Almeida, Carroll Wainwright, Pamela Mishkin, Chong Zhang, Sandhini Agarwal, Katarina Slama, Alex Ray, John Schulman, Jacob Hilton, Fraser Kelton, Luke Miller, Maddie Simens, Amanda Askell, Peter Welinder, Paul~F Christiano, Jan Leike, and Ryan Lowe.
\newblock Training language models to follow instructions with human feedback.
\newblock In S.~Koyejo, S.~Mohamed, A.~Agarwal, D.~Belgrave, K.~Cho, and A.~Oh, editors, {\em Advances in Neural Information Processing Systems}, volume~35, pages 27730--27744. Curran Associates, Inc., 2022.

\bibitem{pan2023automatically}
Liangming Pan, Michael Saxon, Wenda Xu, Deepak Nathani, Xinyi Wang, and William~Yang Wang.
\newblock Automatically correcting large language models: Surveying the landscape of diverse self-correction strategies, 2023.

\bibitem{penedo2023refinedweb}
Guilherme Penedo, Quentin Malartic, Daniel Hesslow, Ruxandra Cojocaru, Alessandro Cappelli, Hamza Alobeidli, Baptiste Pannier, Ebtesam Almazrouei, and Julien Launay.
\newblock The refinedweb dataset for falcon llm: Outperforming curated corpora with web data, and web data only, 2023.

\bibitem{peng2023instruction}
Baolin Peng, Chunyuan Li, Pengcheng He, Michel Galley, and Jianfeng Gao.
\newblock Instruction tuning with gpt-4.
\newblock {\em arXiv preprint arXiv:2304.03277}, 2023.

\bibitem{reimers-gurevych-2019-sentence}
Nils Reimers and Iryna Gurevych.
\newblock Sentence-{BERT}: Sentence embeddings using {S}iamese {BERT}-networks.
\newblock In {\em Proceedings of the 2019 Conference on Empirical Methods in Natural Language Processing and the 9th International Joint Conference on Natural Language Processing (EMNLP-IJCNLP)}, pages 3982--3992, Hong Kong, China, November 2019. Association for Computational Linguistics.

\bibitem{Scao2022BLOOMA1}
Teven~Le Scao, Angela Fan, Christopher Akiki, Elizabeth-Jane Pavlick, Suzana Ili'c, Daniel Hesslow, Roman Castagn'e, Alexandra~Sasha Luccioni, Franccois Yvon, Matthias Gall{\'e}, Jonathan Tow, Alexander~M. Rush, Stella~Rose Biderman, Albert Webson, Pawan~Sasanka Ammanamanchi, Thomas Wang, Beno{\^i}t Sagot, Niklas Muennighoff, Albert~Villanova del Moral, Olatunji Ruwase, Rachel Bawden, Stas Bekman, Angelina McMillan-Major, Iz~Beltagy, Huu Nguyen, Lucile Saulnier, Samson Tan, Pedro~Ortiz Suarez, Victor Sanh, Hugo Laurenccon, Yacine Jernite, Julien Launay, Margaret Mitchell, Colin Raffel, Aaron Gokaslan, Adi Simhi, Aitor~Soroa Etxabe, Alham~Fikri Aji, Amit Alfassy, Anna Rogers, Ariel~Kreisberg Nitzav, Canwen Xu, Chenghao Mou, Chris~C. Emezue, Christopher Klamm, Colin Leong, Daniel~Alexander van Strien, David~Ifeoluwa Adelani, Dragomir~R. Radev, Eduardo~Gonz'alez Ponferrada, Efrat Levkovizh, Ethan Kim, Eyal~Bar Natan, Francesco~De Toni, G{\'e}rard Dupont, Germ{\'a}n Kruszewski, Giada Pistilli, Hady ElSahar, Hamza
  Benyamina, Hieu~Trung Tran, Ian Yu, Idris Abdulmumin, Isaac Johnson, Itziar Gonzalez-Dios, Javier de~la Rosa, Jenny Chim, Jesse Dodge, Jian Zhu, Jonathan Chang, Jorg Frohberg, Josephine~L. Tobing, Joydeep Bhattacharjee, Khalid Almubarak, Kimbo Chen, Kyle Lo, Leandro von Werra, Leon Weber, Long Phan, Loubna~Ben Allal, Ludovic Tanguy, Manan Dey, Manuel~Romero Mu{\~n}oz, Maraim Masoud, Mar'ia Grandury, Mario vSavsko, Max Huang, Maximin Coavoux, and Mayank Singh.
\newblock Bloom: A 176b-parameter open-access multilingual language model.
\newblock {\em ArXiv}, abs/2211.05100, 2022.

\bibitem{shu2023exploitability}
Manli Shu, Jiongxiao Wang, Chen Zhu, Jonas Geiping, Chaowei Xiao, and Tom Goldstein.
\newblock On the exploitability of instruction tuning, 2023.

\bibitem{principle}
Zhiqing Sun, Yikang Shen, Qinhong Zhou, Hongxin Zhang, Zhenfang Chen, David Cox, Yiming Yang, and Chuang Gan.
\newblock Principle-driven self-alignment of language models from scratch with minimal human supervision.
\newblock {\em arXiv preprint arXiv:2305.03047}, 2023.

\bibitem{alpaca}
Rohan Taori, Ishaan Gulrajani, Tianyi Zhang, Yann Dubois, Xuechen Li, Carlos Guestrin, Percy Liang, and Tatsunori~B. Hashimoto.
\newblock Stanford alpaca: An instruction-following llama model.
\newblock \url{https://github.com/tatsu-lab/stanford_alpaca}, 2023.

\bibitem{xwin-lm}
Xwin-LM Team.
\newblock Xwin-lm, 9 2023.

\bibitem{touvron2023llama}
Hugo Touvron, Thibaut Lavril, Gautier Izacard, Xavier Martinet, Marie-Anne Lachaux, Timothée Lacroix, Baptiste Rozière, Naman Goyal, Eric Hambro, Faisal Azhar, Aurelien Rodriguez, Armand Joulin, Edouard Grave, and Guillaume Lample.
\newblock Llama: Open and efficient foundation language models, 2023.

\bibitem{touvron2023llama2}
Hugo Touvron, Louis Martin, Kevin Stone, Peter Albert, Amjad Almahairi, Yasmine Babaei, Nikolay Bashlykov, Soumya Batra, Prajjwal Bhargava, Shruti Bhosale, Dan Bikel, Lukas Blecher, Cristian~Canton Ferrer, Moya Chen, Guillem Cucurull, David Esiobu, Jude Fernandes, Jeremy Fu, Wenyin Fu, Brian Fuller, Cynthia Gao, Vedanuj Goswami, Naman Goyal, Anthony Hartshorn, Saghar Hosseini, Rui Hou, Hakan Inan, Marcin Kardas, Viktor Kerkez, Madian Khabsa, Isabel Kloumann, Artem Korenev, Punit~Singh Koura, Marie-Anne Lachaux, Thibaut Lavril, Jenya Lee, Diana Liskovich, Yinghai Lu, Yuning Mao, Xavier Martinet, Todor Mihaylov, Pushkar Mishra, Igor Molybog, Yixin Nie, Andrew Poulton, Jeremy Reizenstein, Rashi Rungta, Kalyan Saladi, Alan Schelten, Ruan Silva, Eric~Michael Smith, Ranjan Subramanian, Xiaoqing~Ellen Tan, Binh Tang, Ross Taylor, Adina Williams, Jian~Xiang Kuan, Puxin Xu, Zheng Yan, Iliyan Zarov, Yuchen Zhang, Angela Fan, Melanie Kambadur, Sharan Narang, Aurelien Rodriguez, Robert Stojnic, Sergey Edunov, and Thomas
  Scialom.
\newblock Llama 2: Open foundation and fine-tuned chat models, 2023.

\bibitem{vu2023koala}
Thuy-Trang Vu, Xuanli He, Gholamreza Haffari, and Ehsan Shareghi.
\newblock Koala: An index for quantifying overlaps with pre-training corpora, 2023.

\bibitem{wang2023large}
Peiyi Wang, Lei Li, Liang Chen, Dawei Zhu, Binghuai Lin, Yunbo Cao, Qi~Liu, Tianyu Liu, and Zhifang Sui.
\newblock Large language models are not fair evaluators, 2023.

\bibitem{wang-etal-2023-self-instruct}
Yizhong Wang, Yeganeh Kordi, Swaroop Mishra, Alisa Liu, Noah~A. Smith, Daniel Khashabi, and Hannaneh Hajishirzi.
\newblock Self-instruct: Aligning language models with self-generated instructions.
\newblock In {\em Proceedings of the 61st Annual Meeting of the Association for Computational Linguistics (Volume 1: Long Papers)}, pages 13484--13508, Toronto, Canada, July 2023. Association for Computational Linguistics.

\bibitem{wei2022finetuned}
Jason Wei, Maarten Bosma, Vincent Zhao, Kelvin Guu, Adams~Wei Yu, Brian Lester, Nan Du, Andrew~M. Dai, and Quoc~V Le.
\newblock Finetuned language models are zero-shot learners.
\newblock In {\em International Conference on Learning Representations}, 2022.

\bibitem{wei2021finetuned}
Jason Wei, Maarten Bosma, Vincent~Y Zhao, Kelvin Guu, Adams~Wei Yu, Brian Lester, Nan Du, Andrew~M Dai, and Quoc~V Le.
\newblock Finetuned language models are zero-shot learners.
\newblock {\em arXiv preprint arXiv:2109.01652}, 2021.

\bibitem{wei2023chainofthought}
Jason Wei, Xuezhi Wang, Dale Schuurmans, Maarten Bosma, Brian Ichter, Fei Xia, Ed~Chi, Quoc Le, and Denny Zhou.
\newblock Chain-of-thought prompting elicits reasoning in large language models, 2023.

\bibitem{xu2023wizardlm}
Can Xu, Qingfeng Sun, Kai Zheng, Xiubo Geng, Pu~Zhao, Jiazhan Feng, Chongyang Tao, and Daxin Jiang.
\newblock Wizardlm: Empowering large language models to follow complex instructions, 2023.

\bibitem{xu2023baize}
Canwen Xu, Daya Guo, Nan Duan, and Julian McAuley.
\newblock Baize: An open-source chat model with parameter-efficient tuning on self-chat data.
\newblock {\em arXiv preprint arXiv:2304.01196}, 2023.

\bibitem{yan2023virtual}
Jun Yan, Vikas Yadav, Shiyang Li, Lichang Chen, Zheng Tang, Hai Wang, Vijay Srinivasan, Xiang Ren, and Hongxia Jin.
\newblock Virtual prompt injection for instruction-tuned large language models, 2023.

\bibitem{yang2023harnessing}
Jingfeng Yang, Hongye Jin, Ruixiang Tang, Xiaotian Han, Qizhang Feng, Haoming Jiang, Bing Yin, and Xia Hu.
\newblock Harnessing the power of llms in practice: A survey on chatgpt and beyond, 2023.

\bibitem{yao2023tree}
Shunyu Yao, Dian Yu, Jeffrey Zhao, Izhak Shafran, Thomas~L. Griffiths, Yuan Cao, and Karthik Narasimhan.
\newblock Tree of thoughts: Deliberate problem solving with large language models, 2023.

\bibitem{selfee2023}
Seonghyeon Ye, Yongrae Jo, Doyoung Kim, Sungdong Kim, Hyeonbin Hwang, and Minjoon Seo.
\newblock Selfee: Iterative self-revising llm empowered by self-feedback generation.
\newblock Blog post, May 2023.

\bibitem{zellers-etal-2019-hellaswag}
Rowan Zellers, Ari Holtzman, Yonatan Bisk, Ali Farhadi, and Yejin Choi.
\newblock {H}ella{S}wag: Can a machine really finish your sentence?
\newblock In {\em Proceedings of the 57th Annual Meeting of the Association for Computational Linguistics}, pages 4791--4800, Florence, Italy, July 2019. Association for Computational Linguistics.

\bibitem{zhang2023instruction}
Shengyu Zhang, Linfeng Dong, Xiaoya Li, Sen Zhang, Xiaofei Sun, Shuhe Wang, Jiwei Li, Runyi Hu, Tianwei Zhang, Fei Wu, and Guoyin Wang.
\newblock Instruction tuning for large language models: A survey, 2023.

\bibitem{zhao2023survey}
Wayne~Xin Zhao, Kun Zhou, Junyi Li, Tianyi Tang, Xiaolei Wang, Yupeng Hou, Yingqian Min, Beichen Zhang, Junjie Zhang, Zican Dong, Yifan Du, Chen Yang, Yushuo Chen, Zhipeng Chen, Jinhao Jiang, Ruiyang Ren, Yifan Li, Xinyu Tang, Zikang Liu, Peiyu Liu, Jian-Yun Nie, and Ji-Rong Wen.
\newblock A survey of large language models, 2023.

\bibitem{zheng2023judging}
Lianmin Zheng, Wei-Lin Chiang, Ying Sheng, Siyuan Zhuang, Zhanghao Wu, Yonghao Zhuang, Zi~Lin, Zhuohan Li, Dacheng Li, Eric.~P Xing, Hao Zhang, Joseph~E. Gonzalez, and Ion Stoica.
\newblock Judging llm-as-a-judge with mt-bench and chatbot arena, 2023.

\end{thebibliography}
\bibliographystyle{plain}

\end{document}